\relax
\documentclass[letterpaper]{article} 
\usepackage{aaai20}  
\usepackage{times}  
\usepackage{helvet} 
\usepackage{courier}  
\usepackage[hyphens]{url}  
\usepackage{graphicx} 
\usepackage{graphicx}  
\usepackage{booktabs}
\usepackage{textcomp}
\usepackage{amsmath}
\usepackage{graphicx}
\usepackage{bm}
\usepackage{array}
\usepackage{subfigure}
\usepackage{tikz}
\usepackage{verbatim}
\usepackage{subfigure}
\usepackage{enumitem}
\usepackage[ruled]{algorithm2e} 
\usepackage{natbib}

\SetAlFnt{\small}
\SetAlCapFnt{\small}
\SetAlCapNameFnt{\small}
\SetAlCapHSkip{0pt}
\IncMargin{-\parindent}

\let\cite\citep

\title{Simplify-then-Translate: Automatic Preprocessing for Black-Box Translation}

\author{Sneha Mehta\textsuperscript{\rm 1}, Bahareh Azarnoush\textsuperscript{\rm 2}, Boris Chen\textsuperscript{\rm 2}, Avneesh Saluja\textsuperscript{\rm 2}, \\ \Large\textbf{Vinith Misra}\textsuperscript{\rm 2}\Large\textbf{,} \Large\textbf{Ballav Bihani}\textsuperscript{\rm 2}\Large\textbf{,} \Large\textbf{{}Ritwik Kumar}\textsuperscript{\rm 2} \\ 
\textsuperscript{\rm 1}Department of Computer Science, Virginia Tech, VA \\ 
\textsuperscript{\rm 2}Netflix Inc., CA\\
snehamehta@vt.edu, \{bazarnoush, bchen, asaluja, \\ vmisra, bbihani, ritwikk\}@netflix.com
}
 
\begin{document}

\maketitle

\begin{abstract}
Black-box machine translation systems have proven incredibly useful for a variety of applications yet by design are hard to adapt, tune to a specific domain, or build on top of. 
In this work, we introduce a method to improve such systems via automatic pre-processing (APP) using sentence simplification. 
We first propose a method to automatically generate a large in-domain paraphrase corpus through back-translation with a black-box MT system, which is used to train a paraphrase model that ``simplifies" the original sentence to be more conducive for translation.
The model is used to preprocess source sentences of multiple low-resource language pairs. We show that this preprocessing leads to better translation performance as compared to non-preprocessed source sentences.
We further perform side-by-side human evaluation to verify that translations of the simplified sentences are better than the original ones.
Finally, we provide some guidance on recommended language pairs for generating the simplification model corpora by investigating the relationship between ease of translation of a language pair (as measured by BLEU) and quality of the resulting simplification model from back-translations of this language pair (as measured by SARI), and tie this into the downstream task of low-resource translation. 
\end{abstract}




 
\section{Introduction}
Modern translation systems built on top of a sequence transduction approach \cite{Sutskever2014,Bahdanau2014NeuralMT} have greatly advanced the state and quality of machine translation (MT).
These systems generally rely on the availability of large-scale parallel corpora, and while unsupervised ~\cite{lample2017unsupervised} or semi-supervised \cite{Saluja2014} approaches are a popular area of research, production-grade translation systems still primarily leverage bitexts when training.
Efforts such as WMT\footnote{http://www.statmt.org/wmt18/} provide such corpora for select language pairs, which has enabled neural MT systems to achieve state-of-the-art performance on those pairs.
However, low resource MT (language pairs for which parallel data is scarce) remains a challenge.

In this work, we focus on improving translation quality for low-resource translation (i.e., from English \emph{into} a low-resource language) in the \emph{black-box MT} (BBMT) setting - namely a system which has been trained and tuned {\it a priori} and for which we cannot access the model parameters or training data for fine-tuning or improvements.
Examples of such systems include those provided by commercial vendors e.g., Google Translate\footnote{https://translate.google.com/} or Microsoft Translator\footnote{https://www.bing.com/translator}.
While some provide the option of fine-tuning on domain-specific data under certain conditions, how to improve the performance of such black-box systems on domain-specific translation tasks remains an open question.
We investigate methods to leverage the BBMT system to preprocess input source sentences in a way that preserves meaning and improves translation in the target language.
Specifically, a large-scale parallel corpus for English simplification is obtained by back-translating \cite{sennrich2015improving} the reference translations of several high-resource target languages. 
The resulting parallel corpus is used to train an Automatic Preprocessing model (APP) 
(\S~\ref{sec:app}), which transforms source sentence into a form that preserves meaning while also being easier to translate into a low-resource language.
In effect, the APP model attacks the longstanding problem of handling complex idiomatic and non-compositional phrases \cite{Lin1999,Sag2002}, and simplifies these expressions to more literal, compositional ones that we hypothesize are easier to translate.

We use the APP model to simplify the source sentences of a variety of low-resource language pairs and compare the performance of the black-box MT system on the original and simplified sentences.
Note that only one APP model needs to be trained per source language and this model can be applied to a variety of low-resource language pairs as long as the source language is the same. 
In our study we focus on the domain of conversational language as used in dialogues of TV shows. We picked this domain since here language tends to be colloquial and idiomatic. We empirically show improvement in translation quality in this domain across a variety of low resource target languages (\S\ref{sec:experiments}).
This improvement is further corroborated with side-by-side human evaluations (\S\ref{sec:human_eval}) and evaluating on post-editing efficiency (\S\ref{sec:post_edit}). 
Lastly, we perform an empirical analysis to probe further into which high-resource language pairs should be selected to obtain a good quality simplification corpus for a given language, before discussing connections with related work (\S\ref{sec:related}) and concluding. 
\begin{figure}
  \centering
  \includegraphics[width=0.45\textwidth]{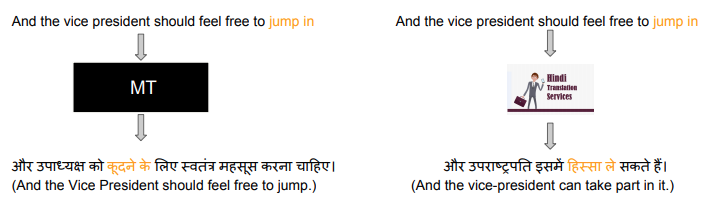}
  \caption{Machine vs. Human Translation
}
\label{mot1}
\end{figure}

\section{APP with Back Translations}\label{sec:app}
Consider the example in Fig. \ref{mot1}.
The source sentence ``The vice president should feel free to jump in" has been translated by Google Translate~\footnote{This specific translation is observed in translate.google.com as of September 5, 2019.} incorrectly to Hindi as ``Vice President should feel free to jump inside".
The system was unable to correctly translate the idiomatic and non-compositional phrase ``jump in", which in this context means ``intervene" or ``get involved", and instead translated it literally. 
An expert human Hindi translator would take these idiomatic expressions into account when generating the Hindi translation. Indeed, when back-translating the reference translation into English we obtain ``The Vice President should feel free to take part" (in the conversation).
Such instances where MT systems incorrectly translate sentences containing phrases, idioms or complex words for low-resources (i.e. with smaller training sets) languages pairs are fairly common.

In other words, the back-translation is different in meaning than the natural source sentence.
This problem is prevalent even when these BBMT systems are fine-tuned on domain-specific data and is exacerbated when dealing with low-resource language pairs, simply because the paucity of data does not allow the MT models to infer the translations of the myriad of phrases and complex words.
For instance, in the example above `jump in' was interpreted compositionally as `jump' $+$ `in'.
To generate better quality or even acceptable translations, it is imperative to simplify such complex sentences into simpler forms while still preserving meaning.

Using automated models to simplify such sentences is a well-studied problem. Though, when it comes to the ultimate task of domain-specific translation, it is not entirely clear what data is best suited to train such simplification models. Open source datasets like WikiLarge ~\cite{zhang2017sentence} or Simple PPDB ~\cite{zhao2018integrating} are good options to explore, but the domain mismatch and dataset size may pose a challenge. In particular, WikiLarge dataset contains 296K sentence pairs of descriptive text while our domain of interest in this study is conversational dialogues from TV Shows and movies. Collecting a large amount of domain-specific simplification data could be prohibitive, forcing one to consider alternatives when constructing their simplification models.


To address this problem we make use of the observation that translating back-translations is easier than translating naturally occurring source sentences, which has been corroborated by numerous studies ~\cite{DBLP:journals/corr/abs-1906-09833,DBLP:journals/corr/abs-1906-08069}. 
Consider a set of around 30K uniformly-sampled sentence pairs from the English-Bulgarian (En-Bg) subtitles corpus appearing on TV shows and movies from a subscription video-on-demand provider\footnote{www.netflix.com}.
 The BLEU score in the natural or direct direction (En $\rightarrow$ Bg) is only 10.20, whereas when following the reverse direction (Bg $\rightarrow$ En $\rightarrow$ Bg) and translating back-translations instead of original source sentences, the BLEU score dramatically improves to 33.39.
 Probing a little deeper, Fig. 2 shows the distributions of sentence-level GLEU scores ~\cite{wu2016google} for two language pairs: English-Bulgarian (right) and English-Hindi (left). 
 We observe the trend that GLEU scores have generally improved when using back-translations. 
 The area where the blue curve dominates the red curve can be considered as the `scope-of-simplification'.


\begin{figure}
\label{fig:before_after}
 \centering
  \subfigure[English-Hindi]{\includegraphics[width=0.23\textwidth]{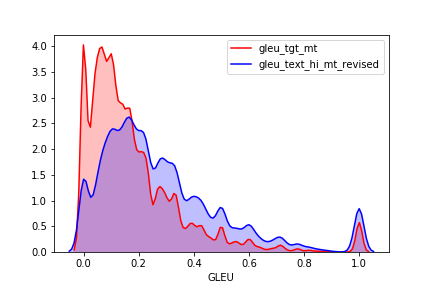}}
  \subfigure[English-Bulgarian]{\includegraphics[width=0.23\textwidth]{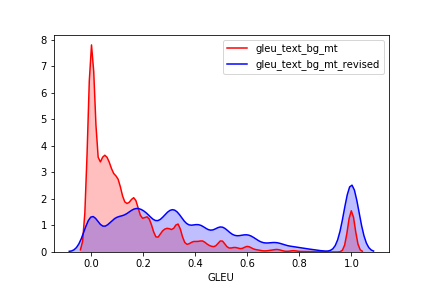}}
 \caption{\small{Sentence-level GLEU for direct translations and back-translations. The red density curve represents the distribution of GLEU scores obtained by the direction translation direction and the blue density curve represents the distribution of GLEU scores obtained by using back-translations.}}
\end{figure}

Thus it seems  that human reference translations when back-translated to the original language (in this case English) are a rich source of simplifications (e.g. ``jump in" is simplified to ``take part").
This observation leads to two immediate corollaries - 1) by back-translating the ground truth human translations to the source language we obtain a (perhaps noisy) simplified version of the original source, and 2) we can learn a function to map the source sentences to their simplified forms by training a sequence-to-sequence (S2S) model from the aforementioned generated parallel corpus.
We term the resulting simplification model an {\bf Automated Preprocessing} (or APP) model.

We formalize our APP model as a S2S model from source sentences in one language to back-translations of ground truth sentences i.e., \textit{translationese} targets in the same language.
The \textit{translationese} targets can be obtained from multiple high-resource language pairs using black-box MT systems and the trained APP model can be applied to a variety of low resource language pairs.
Let $(X^i, Y^i)$ be the $i^\textrm{th}$ training bitext corpus ($X$ is source, $Y$ is target) with source language $s_i$ and target language $t_i$ for $i \in \{1, \dots, M\}$, where $M$ is the number of training language pairs, and let $j \in \{M+1, \dots, M+N\}$ refer to the $N$ test bitext corpora.
Note that for the experiments in this paper, $s_i$ is fixed to English $\forall i \in \{1, \dots, M+N\}$ so we simply refer to it as $s$.
The APP procedure is as follows:
\begin{itemize}
    \item Obtain back-translations of target train sets $Y^i$ for $i=1$ to $M$ to language $s_i$ given by $T^1, T^2, \dots, T^M$ using BBMT models $MT_{t_i \rightarrow s} \forall i$.
    \item Train an APP simplification model $f_{APP}$ on the combined parallel corpus $\bigcup_{i=1}^M \{(X^i,T^i)\}$.
    \item At test time, preprocess the source $X_s^j$ for each test language pair $j$ using the trained APP model to obtain the simplified source $X^{j*}$, where
    \begin{equation}
        X^{j*} = f_{APP}(X^j)
    \end{equation}
    \item Translate the simplified source using the BBMT model for the $j^{th}$ test language pair.
    \begin{equation}
    \hat{Y}^{j*} = MT_{s \rightarrow t_j}(X^{j*})
    \end{equation}
\end{itemize}

APP provides in-domain simplification bitext at scale and from the same BBMT system that we eventually use to translate into low-resource languages, thus providing a more flexible solution than using precompiled simplification corpora.
In the next sections, we compare the performance of BBMT system outputs with and without APP simplifications i.e. $\hat{Y}^j*$ \& $\hat{Y}^j$ respectively. Further, we compare the APP models trained on in-domain vs out-of-domain corpora.

\section{Evaluation}
\label{sec:experiments}
We first compare the in-domain APP model to a S2S model trained on the WikiLarge corpus, by evaluating downstream translation performance on low-resource languages, followed by an evaluation based on human judgements and post-editing efficiency. 
These experiments are conducted on subtitles dataset from a subscription video-on-demand provider.
We also validate the approach on another subtitles dataset and verify that the improvements we see in the first set of experiments generalize to other corpora in the same domain.
In all of our experiments we used the Google Translate BBMT system.

\subsection{Datasets and Metrics}

\subsubsection{FIGS}
This dataset comes from subtitles appearing on 12,301 TV shows and movies from a subscription video-on-demand provider. Titles include but are not limited to: ``How to Get Away with Murder", ``Star Trek: Deep Space Nine", and ``Full Metal Alchemist". We take four high-resource language pairs
namely: English-\textbf{F}rench (1.3 million parallel subtitle events i.e., sentence pairs), English-\textbf{I}talian (1.0M), English-\textbf{G}erman (1.2M) and English-\textbf{S}panish (6.5M).
We collectively refer to this dataset  as FIGS.
We use the APP simplification procedure to obtain  English simplification parallel corpora resulting in 9.5M subtitle events i.e., sentence pairs.
This dataset contains short sentences with an average length of 7.
For evaluation, we pick 7 low-resource language pairs namely:
English-Hungarian (En-Hu), English-Ukrainian (En-Uk), English-Czech (En-Cs),
English-Romanian (En-Ro),
English-Bulgarian (En-Bg),
English-Hindi (En-Hi),
and English-Malay (En-Ms).
The test set statistics for each dataset is given in Table ~\ref{tab:netflix}.
We refer to the APP model trained on this dataset as `FigsAPP'.

\begin{table}[!t]
\small
\caption{\small{FIGS test set statistics }}
\label{tab:netflix}
\center
\begin{tabular}{@{}ll@{}}
\toprule
Language & \#sentences \\ \midrule
En-Hu       & 27,393       \\
En-Uk       & 30,761       \\
En-Cs       & 35,505       \\
En-Ro       & 47,054       \\
En-Bg       & 30,714       \\
En-Hi       & 23,496       \\
En-Ms       & 11,713       \\ \bottomrule
\end{tabular}
\end{table}

\subsubsection{Wikilarge}
The WikiLarge dataset ~\cite{zhang2017sentence} was compiled by using sentence alignments from other Wikipedia-based datasets ~\cite{Zhu:2010:MTT:1873781.1873933,Woodsend:2011:LSS:2145432.2145480,kauchak2013improving}, and contains 296K instances of 1-to-1 and 1-to-many alignments.
This is a widely used benchmark for test simplification tasks.
The train split contains 296K sentence pairs and the validation split contains 992 sentence pairs.
We call the simplification model trained on this dataset as `WikiAPP'.

\subsubsection{Open Subtitles}
The Open Subtitles dataset ~\cite{lison2016opensubtitles2016} is a collection of translated movie subtitles obtained from opensubtitles.org\footnote{http://www.opensubtitles.org/ }.
It contains 62 languages and 1,782 bitexts. We first train two MT models using two high resource language pairs (Es-En and Ru-En) using the Transformer architecture ~\cite{Vaswani2017AttentionIA}. Then using the MT models above we train two APP models obtained from the same language pairs English-Spanish (En-Es) and English-Russian (En-Ru). We sample 5M sentence pairs each for training MT and APP models from the corresponding Open Subtitles corpora and filter out short ($length < 3$) and long ($length > 50$) sentence pairs. Note that training sets for both MT and APP models are disjoint.
For evaluation, we pick the following six language pairs: three randomly picked pairs English-Armenian (En-Hy), English-Ukrainian(En-Uk) and English-Bulgarian (En-Bg) and three pairs in which the target language is similar to Spanish including English-Catalan (En-Ca), English-Portuguese (En-Pt) and English-Romanian (En-Ro).
We sample 50,000 pairs from the low-resource test bitexts and filter out pairs with length less than 3 and greater than 50.
We call the APP model obtained from En-Es dataset as `OSEsAPP' and from En-Ru dataset as `OSRuAPP'.

\subsubsection{Turk and PWKP}
To test the performance of APP models obtained from different language pairs (~\S~\ref{sec:ease}) we pick two open source simplification datasets.
The first dataset is the Turk ~\cite{xu2016optimizing} dataset which contains 1-to-1 alignments focused on paraphrasing transformations, and multiple (8) simplification references per original sentence (collected through Amazon Mechanical Turk).
To evaluate the performance on this dataset we use the SARI metric they introduced.
The next dataset we use is the test set of the PWKP dataset ~\cite{Zhu:2010:MTT:1873781.1873933}. This dataset contains only 1-to-1 mapping between source and reference and hence we use the BLEU metric to evaluate the simplification performance.
  
\subsubsection{Metrics}
We evaluate the translation performance of the BBMT system using the commonly used BLEU metric ~\cite{papineni2002bleu}.
For subtitle generation, expert linguists post-edit subtitles at the event or dialog level, hence it is useful to look at the impact that simplification brings at the sentence-level, motivating the choice of sentence-level GLEU ~\cite{wu2016google}, which has been shown to be better correlated with human judgements at the sentence-level as compared to sentence-level BLEU. Furthermore, we compute the normalized edit (Levenshtein) distance between a translation output and human reference translation also known as translation error rate (TER), which has been shown to correlate well with the amount of post-editing effort required by a human \cite{Snover2006}. This metric provides yet another way to evaluate the quality of translation and completes our comprehensive suite of metrics.

\subsection{Implementation}
For training the APP simplification model we use the Transformer architecture ~\cite{Vaswani2017AttentionIA} through the tensor2tensor\footnote{https://github.com/tensorflow/tensor2tensor} library.
We also evaluate BLEU using the implementation in that library and report the uncased version of the metric. For computing the TER score we use the implementation provided by ~\cite{Snover2006}~\footnote{http://www.cs.umd.edu/~snover/tercom/}.

All experiments are based on the transformer base architecture with 6 blocks in the encoder and decoder.
We use the same hyper-parameters for all experiments, i.e., word representations of size 512 and feed-forward layers with inner dimension 4096.
Dropout is set to 0.2 and we use 8 attention heads. Models are optimized with Adam ~\cite{kingma2014adam} using $\beta_1 = 0.9$, $\beta_2 = 0.98$, and $\epsilon = 1e-9$, with the same learning rate schedule as ~\citet{Vaswani2017AttentionIA}.
We use 50,000 warmup steps. All models use label smoothing of 0.1 with a uniform prior distribution over the vocabulary.
We run all experiments using machines with 4 Nvidia V100 GPUs.
We use a sub-word vocabulary of size 32K implemented using the word-piece algorithm ~\cite{sennrich2015improving} to deal with out-of-vocabulary words and the open vocabulary problem in S2S language models.

\section{Results}
\begin{figure}
  \centering
  \includegraphics[width=0.45\textwidth]{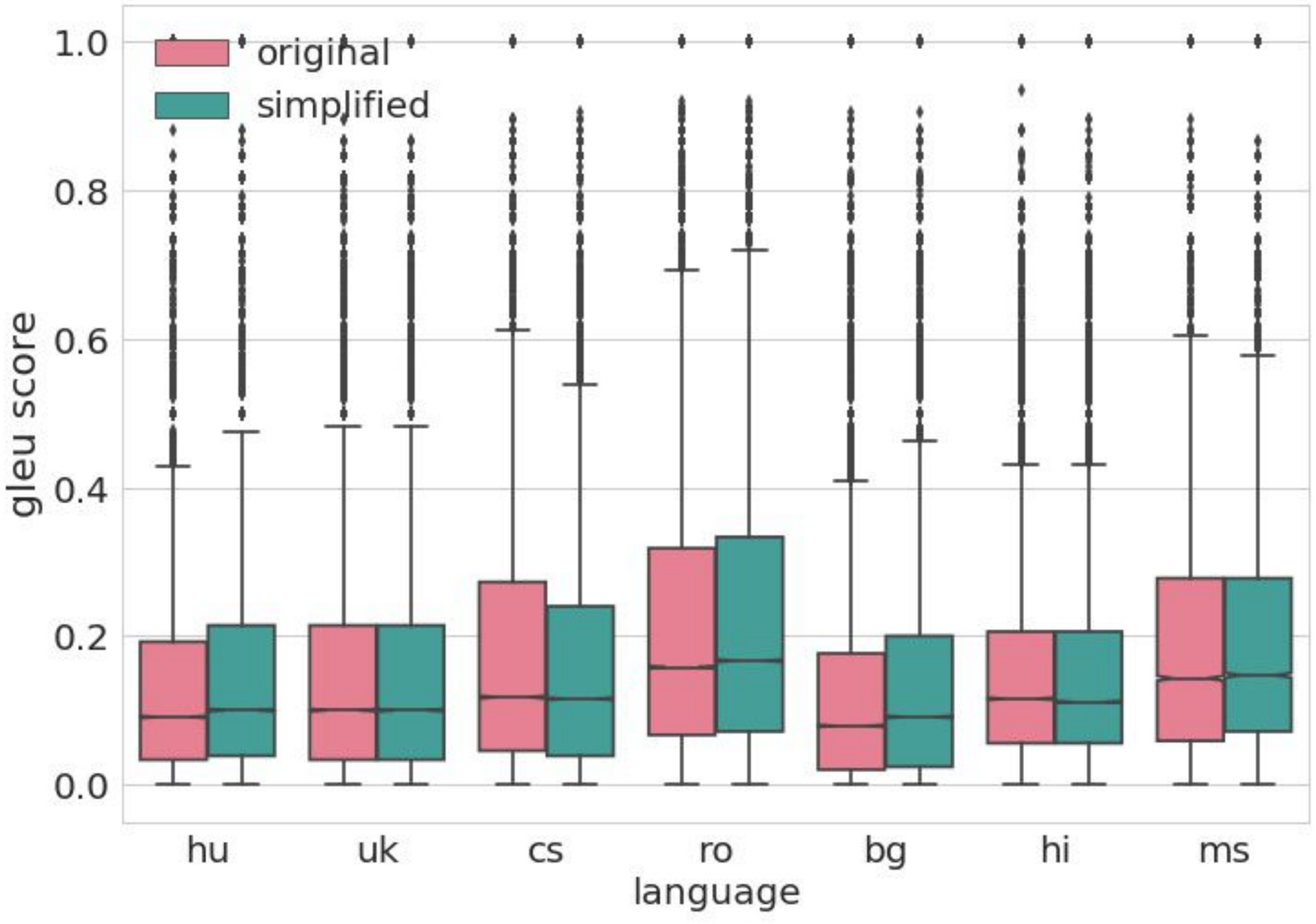}
  \caption{Sentence-level GLEU scores
}
\label{gleu}
\end{figure}

Table ~\ref{tab:nflx_vs_wiki} compares the performance of APP models trained on FIGS (in-domain) dataset and the WikiLarge (out-of-domain) datasets on the FIGS test set using the BBMT system. 
The values in columns 1-3 indicate the BLEU score after translating the original sentence and after simplifying using the FigsAPP and WikiAPP models respectively.
There is uniform improvement across all languages when using the FigsAPP model, ranging from 3.5\% (relative) for En-Uk to 11.6\% for En-Bg. 
On the other hand,  performance degrades significantly on all languages when simplified using a model trained on the WikiLarge dataset.
Fig. ~\ref{gleu} shows the distribution of sentence-level GLEU for each target language in the FIGS test set. Mean GLEU increases for En-Hu, En-Ro, En-Ms and En-Bg. 

Table ~\ref{tab:os_bleu} shows the results of APP on the Open Subtitles dataset. It can be noted that the performance improves for Catalan (ca) and Portuguese (pt) which are languages similar to the language used for training the simplification corpus. Additionally for Bulgarian an improvement of 4.1\% can be observed. Moreover, the performance of OSEsAPP is better than the performance of OSRuAPP. This can be attributed to the fact that En-Ru is a harder language pair to translate than En-Es and hence the simplification model obtained from En-Ru is of worse quality than En-Es. We further elaborate this point in section \S~\ref{sec:ease}.

\begin{table}[!t]
\small
\caption{\small{In-domain vs out-of-domain simplification performance. It is evident that APP models trained on an out-of-domain simplification corpus (WikiLarge) degrades performance, whereas in-domain simplification corpora (FIGS) boosts performance.}}
\label{tab:nflx_vs_wiki}
\centering
\begin{tabular}{@{}llll@{}}
\toprule
language pair                                                       & original & FigsAPP & WikiApp \\ \midrule
En-Hu                                                             & 17.69         & \textbf{18.92}       & 11.86       \\
En-Uk                                                             & 17.57         & \textbf{18.18}       & 12.51       \\
En-Cs                                                             & 21.62         & \textbf{22.35}       & 16.71       \\
En-Ro                                                             & 26.08         & \textbf{27.98}       & 21.56       \\
En-Bg                                                             & 14.9          & \textbf{16.63}       & 12.39       \\
En-Hi                                                             & 14.45         & \textbf{15.53}       & 11.39       \\
En-Ms                                                             & 19.14         & \textbf{20.37}       & 13.00       \\ \bottomrule
\end{tabular}

\end{table}

\begin{table}[!t]
\small
\caption{\small{Translation performance (BLEU) before and after of OSEsAPP and OSRuAPP models on six low resource target language pairs from the Open Subtitles corpus.}}
\label{tab:os_bleu}
\centering
\begin{tabular}{@{}llll@{}}
\toprule
language pair & original & OSEsAPP & OSRuAPP \\ \midrule
En-Ca       & 27.25 & \textbf{27.84}  & 23.36 \\
En-Hu       & \textbf{7.05}          & 6.28  & 5.79     \\
En-Pt       & 25.11         & \textbf{25.5}  &  22.28    \\
En-Ro       & \textbf{26.18}         & 25.03  &  22.40   \\
En-Uk       & 11.73         &  \textbf{11.77} &  11.61    \\
En-Bg       & 23.71         & \textbf{24.68} & 23.90      \\ \bottomrule
\end{tabular}
\end{table}


\subsection{Post-Editing Efficiency}
\label{sec:post_edit}

\begin{table}[!ht]
\small
\caption{\small{Translation Edit Rate (TER) score of translations before and after applying the APP simplification for test language-pairs from the FIGS dataset. `original' and `simple' columns show the TER for translations before and after APP where the last column indicates the percentage decrease in TER.}}
\label{tab:ter_scores}
\centering
\begin{tabular}{@{}llll@{}}
\toprule
language & original & simple & \%$\Delta$\\ \midrule
En-Hu       & 0.86     & 0.80 &  -7\% \\
En-Uk       & 0.74     & 0.70 & -5.4\% \\
En-Cs       & 0.78     & 0.77 & -1.3\% \\
En-Ro       & 0.72     & 0.67 & -6.9\% \\
En-Bg       & 0.83     & 0.77 & -7.3\% \\
En-Hi       & 0.62     & 0.60 & -3.3\% \\
En-Ms       & 0.76     & 0.72 & -5.3\% \\ \bottomrule
\end{tabular}
\end{table}


We also observe that TER decreases for all languages, which is intuitive to understand because the APP simplification brings the sentences closer to their literal human translation. Table ~\ref{tab:ter_scores} shows TER score for the FIGS test corpora for translating into seven low resource languages, before and after applying the APP simplification. As can be seen, TER decreases for all languages after simplification with a reduction of 6.9\%, 6.9\% and 7.2\% for target languages Hungarian, Romanian and Bulgarian respectively. The reduction in TER correspondingly translates to a reduction in post-editing effort required by translators using the BBMT system as an assistive tool.

\subsection{Human Evaluation}
\label{sec:human_eval}
Simplified sentences with worse GLEU than their baseline non-simplified counterparts  might not necessarily be of worse quality; rather, they may just be phrased differently than the reference sentence.
We thus perform a
\begin{figure}
  \centering
  \includegraphics[width=0.5\textwidth]{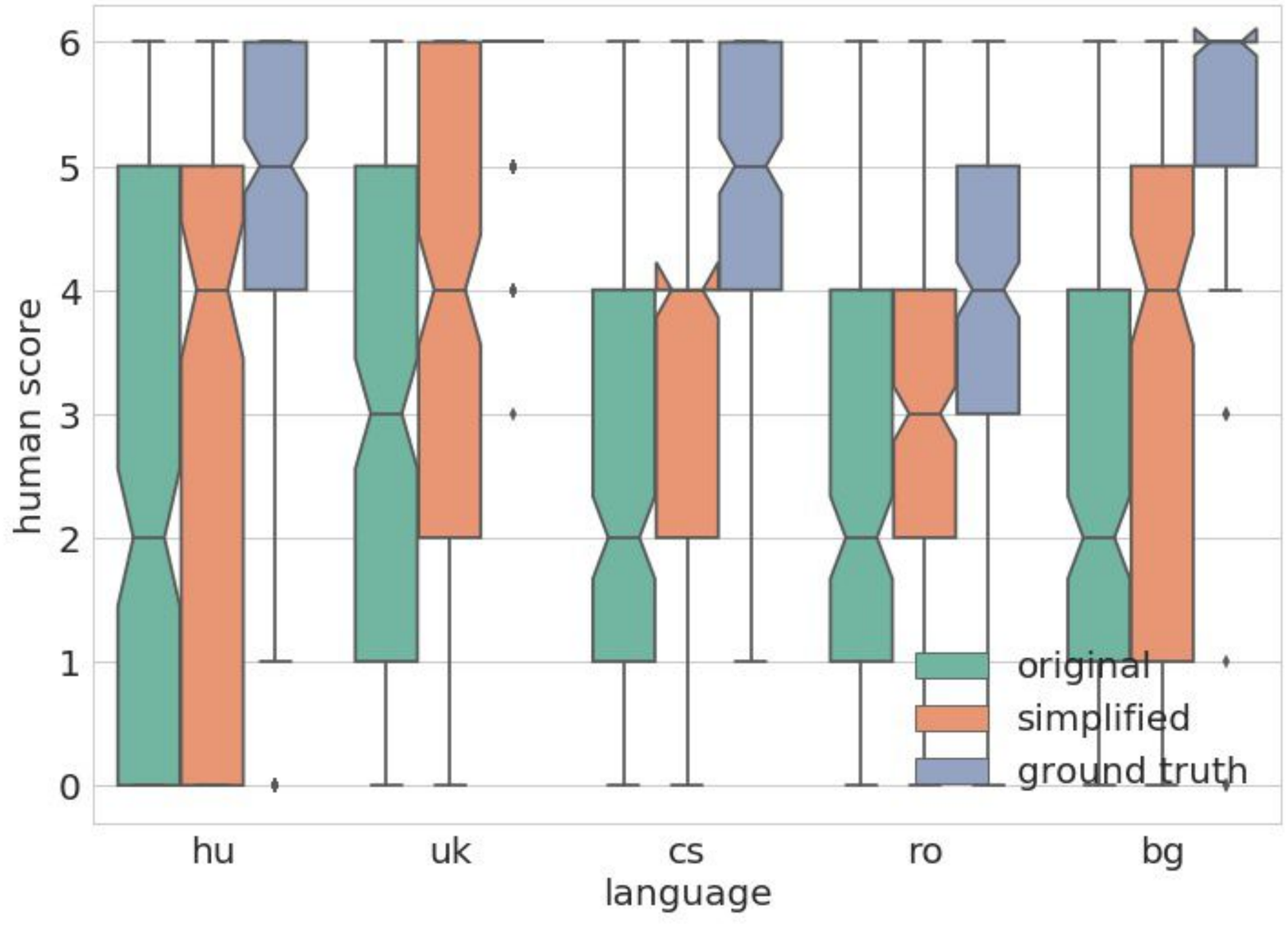}
  \caption{Human Evaluation Results
}
\label{human}
\end{figure}
side-by-side human evaluation to verify if APP-simplified translations improve MT quality, which allows us to assess via human evaluation these supposedly worse translations and validate translations with GLEU improvements at the same time. 
For this purpose, we restricted evaluation to five languages from the FIGS test set (hu, uk, cs, ro, and bg) and sampled 100 sentences from the fraction of sentences for which $\Delta GLEU > 0.4$ and 100 sentences from the sentences for which $\Delta GLEU < 0$, where $\Delta GLEU$ for one sentence pair (x, y) is defined as; 
\begin{equation}
\Delta GLEU=GLEU(MT(x^*), y) - GLEU(MT(x), y)  
\end{equation}
and $x^*$ is the simplified source sentence.
For each language we show the source sentence, the BBMT output of the original sentence, the BBMT output of the simplified sentence, and the ground truth human translation per language side-by-side to expert linguists.
We ask them to rate the quality of the three translations (original BBMT, simplified BBMT and human translation) according to the scale used by ~\citet{wu2016google} described in Table ~\ref{tab:human_eval}.
Since translation is generally easier for shorter sentences, in order to get a representative sample of challenging sentences we only selected sentences with more than 4 tokens for this study. 

\begin{table}[!ht]
\small
\caption{\small{Human evaluation ratings and their description}}
\label{tab:human_eval}
\begin{tabular}{@{}ll@{}}
\toprule
Rating & Description                                                                                                                                                         \\ \midrule
0      & completely nonsense translation                                                                                                                                     \\ \midrule
2      & \begin{tabular}[c]{@{}l@{}}the sentence preserves some of the meaning of the \\ source sentence but misses significant parts\end{tabular}                            \\ \midrule
4      & \begin{tabular}[c]{@{}l@{}}the sentence retains most of the meaning of the source \\ sentence, but may have some grammar mistakes\end{tabular}                      \\ \midrule
6      & \begin{tabular}[c]{@{}l@{}}perfect translation: the meaning of the translation is \\ completely  consistent with the source, and the grammar \\ is correct\end{tabular} \\ \bottomrule
\end{tabular} 
\end{table}

Fig ~\ref{human} displays results from the human evaluation, with corresponding values stated in Table \ref{tab:human_eval_stat}.
The green box represents scores from translations of original sentences, the orange box represents scores from translations from simplified sentences and the blue box represents the ground truth human translation. 
As expected the human translation has the highest score. Also worth noting is the jump in score intervals between original and simplified translations.
For Hungarian, Czech, and Bulgarian median scores jump from $2 \rightarrow 4$, with improvements for Ukrainian ($3 \rightarrow 4$) and Romanian ($2 \rightarrow 3$) as well.
Thus at least for these languages we can conclude that APP simplification results in improved translation output, as determined by expert human translators. 
This concurs with the initial observation e.g., for Bulgarian in Fig. 2, where the number of sentences with score of 0.0 are squashed and their scores shifted higher.
This means simplification can improve upon erroneous or bad-quality translations.
Table ~\ref{tab:human_eval_stat} shows the mean human scores for original, simplified and reference translations as well as percentage of sentences for which human score improved, worsened and remained the same after simplification for all 200 sentences per language.

\begin{table}[!ht]
\small
\caption{\small{Human Evaluation statistics for the FIGS test language pairs.  `Original Mean', `Simple Mean' and `Human Mean' represent the mean human scores before, after APP and ground truth human scores respectively. The next three columns indicates the percentage of samples with improved (\% +ve), worse (\% -ve), and same (\% same) performance after simplification. }}
\label{tab:human_eval_stat}
\begin{tabular}{@{}lllllll@{}}
\toprule
lang & \begin{tabular}[c]{@{}l@{}}Original\\ Mean\end{tabular} & \begin{tabular}[c]{@{}l@{}}Simple\\ Mean\end{tabular} & \begin{tabular}[c]{@{}l@{}}Human\\ Mean\end{tabular} & \% +ve  & \% -ve  & \% same \\ \midrule
Hu   & 2.52                                                    & 3.11                                                  & 4.45                                                 & 38.5\% & 18.5\% & 43\%    \\
Uk   & 3.02                                                    & 3.61                                                  & 5.8                                                  & 43\%    & 15\%    & 42\%    \\
Cs   & 2.43                                                    & 3.12                                                  & 4.77                                                 & 49\%    & 16.5\%  & 34.5\%  \\
Ro   & 2.56                                                    & 3.0                                                   & 3.91                                                 & 41.5\% & 25.5\%  & 33\%    \\
Bg   & 2.34                                                    & 3.33                                                  & 5.365                                                & 51\%    & 22\%    & 27\%    \\ \bottomrule
\end{tabular}
\end{table}

\subsection{Ease of Translation vs Simplification Quality} \label{sec:ease}
We further seek to investigate the relationship between the translation language pair used to generate the simplification corpus and the quality of the resulting simplification model.
To this end, we train translation systems on two language pairs: English-Spanish (En-Es) and English-Russian (En-Ru). 
The En-Es pair is an \textit{easier} language pair to translate than En-Ru as reflected by the BLEU scores of the SOTA MT systems on these pairs.
We pick an En-Es system trained to reach a BLEU of 42.7 and an En-Ru system trained to a BLEU of 34.23. We hypothesize that the APP models resulting from an easier language pair will be of better quality because it is easier to generate a good quality parallel simplification corpus.
To test this hypothesis, we train simplification models obtained from the En-Es and En-Ru translation pairs and test the simplification performance on two standard simplification test sets.

\begin{table}[]
\caption{\small{Simplification performance of APP models trained on corpora generated by En-Es and En-Ru datasets on Turk and PWKP open source datasets.}}
\label{tab:turk}
\centering
\begin{tabular}{@{}llll@{}}
\toprule
(Metric) Dataset & Wiki & En-Es & En-Ru \\ \midrule
(SARI) Turk          & 31.9 & 26.7  & 23.2  \\
(BLEU) PWKP          & 53.5 & 32.7  & 23.9 \\ \bottomrule
\end{tabular}
\end{table}

Table ~\ref{tab:turk} presents the results of this experiment, specifically the performance of the simplification models trained via automatically-generated simplification corpora obtained from the En-Es and En-Ru OpenSubtitles models.
We also test the performance of an in-domain simplification model trained on the WikiLarge dataset. As expected, the performance of the in-domain model exceeds the performance of the models trained on En-Es and En-Ru datasets. It is also interesting to note that the En-Es model outperforms the En-Ru models on both datasets, underlining our hypothesis that easier language pairs result in better APP models. This idea can be used to inform the high-resource language pair(s) to pick for training an APP model for a target language pair. For instance, to simplify English it would be better to pick a high-resource pair which is easier to translate from English, e.g. Spanish. While simplifying Catalan(ca), it would be good to pick a translation pair like Catalan-Spanish which would be easier to translate than Catalan-English.

\subsection{Qualitative Analysis}
\begin{table}[]
\small
\caption{\small{Positive and negative qualitative examples of simplifications brought about by APP. }}
\label{tab:qual_simple}

\begin{tabular}{@{}ll@{}}
\toprule
\begin{tabular}[c]{@{}l@{}}Original:\\ Simple:\end{tabular} & \begin{tabular}[c]{@{}l@{}}I still, think \textbf{you're nuts}, but not as nuts as I thought\\ I still think \textbf{you're crazy}, but not as crazy as I thought\end{tabular} \\ \midrule
\begin{tabular}[c]{@{}l@{}}Original:\\ Simple:\end{tabular} & \begin{tabular}[c]{@{}l@{}}in a town \textbf{only five miles} from Kabul.\\ in a city \textbf{eight kilometers} from Kabul.\end{tabular}                                        \\ \midrule
\begin{tabular}[c]{@{}l@{}}Original:\\ \\ Simple:\end{tabular} & \begin{tabular}[c]{@{}l@{}} This case is \textbf{so far over your head}, it'd make your nose \\ bleed.\\ This case is \textbf{so complicated} that it would bleed your nose.
\end{tabular}                                        \\ \midrule
\begin{tabular}[c]{@{}l@{}}Original:\\ Simple:\end{tabular} & \begin{tabular}[c]{@{}l@{}} When I was \textbf{marooned} here, my first meal was a pheasant.\\ When I was \textbf{stranded} here, my first meal was a pheasant. \end{tabular} \\\midrule
\begin{tabular}[c]{@{}l@{}}Original:\\ Simple:\end{tabular} & \begin{tabular}[c]{@{}l@{}} \textit{She} jumped from the window of Room 180.\\ \textit{He} jumped out of the window of Room 180. \end{tabular} \\
\bottomrule
\end{tabular}
\end{table}
Here we provide qualitative examples of simplifications generated by the APP approach and how it helps in improving BBMT performance. Table ~\ref{tab:qual_simple} shows some example simplifications from the FIGS test datasets. Phrases highlighted in bold were converted to `simpler' phrases. In the first example ``you're nuts" was replaced by a non-idiomatic/simpler phrase ``you're crazy". In the second example, distance of five miles was almost accurately converted from imperial to the metric system whereas in the third example non-compositional phrase ``so far over your head" was translated into the compositional phrase ``so complicated". In the next example, an infrequent word ``marooned'' was replaced by its more frequent counterpart ``stranded''. Finally, the last example shows the kinds of errors introduced by the APP model where it occasionally replaces pronouns like `it', `she' by `he'.

Table ~\ref{tab:qual_mt} gives examples of how APP simplifications can help the BBMT systems make fewer errors. 
 The first example shows a sample from English-Romanian translation pair. Direct translation of the source makes the BBMT system incorrectly translate ``fixating" as ``fixing" (as observed in the backtranslation of the BBMT output) where as simplifying ``fixating on" as ``thinking about" produces a more meaningful translation. Similarly in the second example from the English-Catalan pair of the Open Subtitles corpus, APP replaces the colloquial word `swell' by its meaning `great' and hence results in a translation that is identical to the reference.
 
 \begin{table}[!t]
 \small
\caption{\small{Qualitative examples of how APP simplification can help mitigate BBMT errors on the FIGS and Open Subtitles datasets. Here x is the input source. }}
\label{tab:qual_mt}
\centering
\begin{tabular}{@{}ll@{}}
\toprule
\begin{tabular}[c]{@{}l@{}}x:\\ BBMT(x):\\ APP(x):\\ BBMT(APP(x)):\\ Reference:\end{tabular}    & \begin{tabular}[c]{@{}l@{}}If only we can \textbf{stop fixating} on the days\\ Dacă numai putem opri fixarea pe zile\\ If only we could \textbf{stop thinking} about the days\\ Dacă am putea să nu ne mai gândim la zile \\ Trebuie să nu ne mai gândim la zile \end{tabular} \\ \midrule
\begin{tabular}[c]{@{}l@{}}x:\\ BBMT(x):\\ APP(x):\\ BBMT(APP(x)):\\ Reference:\end{tabular} & \begin{tabular}[c]{@{}l@{}} Another \textbf{swell} party, Jay.\\ Una altra festa de l’onatge, Jay. \\ Another \textbf{great} party, Jay. \\ Una altra gran festa, Jay. \\ Una altra gran festa, Jay.\end{tabular}                                                                                 \\ \bottomrule
\end{tabular}
\end{table}

\section{Related Work}
\label{sec:related}
Automatic text simplification (ATS) systems aim to transform original texts into their lexically and syntactically simpler variants.
In theory, they could also simplify texts at the discourse level, but most systems still operate only on the sentence level.
The motivation for building the first ATS systems was to improve the performance of machine translation systems and other text processing tasks, e.g. parsing, information retrieval, and summarization ~\cite{chandrasekar1996motivations}.
It was argued that simplified
sentences (which have simpler sentential structures and reduced ambiguity) could lead to improvements in the quality of machine translation ~\cite{chandrasekar1996motivations}
A large body of work, since then has investigated text simplification for machine translation and found that this approach can improve fluency of the translation output and reduce technical post-editing effort ~\cite{vstajner2016can}.

Researchers have attempted to build simplification systems for different languages such as English, Spanish, ~\cite{Saggion:2015:MSI:2775084.2738046}, and Portuguese ~\cite{Aluisio:2008:TBP:1410140.1410191}.
\citet{Wubben:2012:SSM:2390524.2390660} use phrase-based machine translation for sentence simplification based on the PWKP dataset ~\cite{Zhu:2010:MTT:1873781.1873933}. However, these systems are modular, rule-based ~\cite{poornima2011rule}, limited by data or language specific.

End-to-end simplification which is more similar to our work also has been studied by applying RNN-based sequence-to-sequence models to the PWKP dataset or transformer based models that are integrated with paraphrase rules ~\cite{zhao2018integrating} and trained on English to English parallel simplification corpora and are current state-of-the-art. These methods are limited by the availability of parallel simplification corpora and especially the ones that can adapt to new domains.
We propose a general framework that can be used to collect large-scale data for any language to train in-domain end-to-end data-driven lexical simplification systems.

Our work capitalizes on the observation that synthetically-generated source sentences resulting from reversing the translation direction on a parallel corpus yield better translations. 
These ``back-translations" ~\cite{sennrich2015neural,poncelas2018investigating} can augment relatively scarce parallel data with by translating the plentiful target monolingual data to the source language. 

Various methods have been explored to improve low-resource translation. ~\cite{zoph2016transfer} transfer parameters from an MT model trained on a high-resource language pair to low-resource language pairs and observe performance improvement. To improve performance on spoken language domain, researchers have finetuned state-of-the-art models trained on domains in which data is more abundant ~\cite{luong2015stanford} whereas others have used data augmentation techniques ~\cite{fadaee2017data} to bring improvements in low-resource translation. Above approaches assume access to the underlying MT system whereas we assume a black-box scenario.

\section{Conclusion}
In this work we introduced a framework for generating a large-scale parallel corpus for sentence simplification, and demonstrated how the corpus can be used to improve the performance of black-box MT systems (especially on low-resource language pairs) and increase the post-editing efficiency at the subtitle-event i.e., sentence level. Moreover, we perform thorough empirical analysis to give insights into language pairs to select for simplifying a given language. Our results suggest that easier a language pair to translate, the better the simplification model that will result.

It should be noted that even though this work mainly focuses on simplification of English, our method is general and can be used to automatically generate simplification parallel corpora and thus data-driven simplification models using state-of-the-art architectures for any given language.
Moreover, it accommodates collecting multiple reference simplifications for a given source sentence by leveraging open-source multilingual corpora. Using the insight that translating multiword expressions and non-compositional phrases is hard and simplifying these expressions before translating helps, our work merges two important sub-fields of NLP (machine translation and sentence simplification) and paves the path for future research in both of these fields.



\bibliographystyle{aaai}
\bibliography{main}

\begin{thebibliography}{}

\bibitem[\protect\citeauthoryear{Alu\'{\i}sio \bgroup et al\mbox.\egroup
  }{2008}]{Aluisio:2008:TBP:1410140.1410191}
Alu\'{\i}sio, S.~M.; Specia, L.; Pardo, T.~A.; Maziero, E.~G.; and Fortes,
  R.~P.
\newblock 2008.
\newblock Towards brazilian portuguese automatic text simplification systems.
\newblock In {\em Proceedings of the Eighth ACM Symposium on Document
  Engineering}, DocEng '08,  240--248.
\newblock New York, NY, USA: ACM.

\bibitem[\protect\citeauthoryear{Bahdanau, Cho, and
  Bengio}{2014}]{Bahdanau2014NeuralMT}
Bahdanau, D.; Cho, K.; and Bengio, Y.
\newblock 2014.
\newblock Neural machine translation by jointly learning to align and
  translate.
\newblock {\em CoRR} abs/1409.0473.

\bibitem[\protect\citeauthoryear{Chandrasekar, Doran, and
  Srinivas}{1996}]{chandrasekar1996motivations}
Chandrasekar, R.; Doran, C.; and Srinivas, B.
\newblock 1996.
\newblock Motivations and methods for text simplification.
\newblock In {\em Proceedings of the 16th conference on Computational
  linguistics-Volume 2},  1041--1044.
\newblock Association for Computational Linguistics.

\bibitem[\protect\citeauthoryear{Fadaee, Bisazza, and
  Monz}{2017}]{fadaee2017data}
Fadaee, M.; Bisazza, A.; and Monz, C.
\newblock 2017.
\newblock Data augmentation for low-resource neural machine translation.
\newblock In {\em ACL},  567--573.
\newblock Vancouver, Canada: Association for Computational Linguistics.

\bibitem[\protect\citeauthoryear{Graham, Haddow, and
  Koehn}{2019}]{DBLP:journals/corr/abs-1906-09833}
Graham, Y.; Haddow, B.; and Koehn, P.
\newblock 2019.
\newblock Translationese in machine translation evaluation.
\newblock {\em CoRR} abs/1906.09833.

\bibitem[\protect\citeauthoryear{Kauchak}{2013}]{kauchak2013improving}
Kauchak, D.
\newblock 2013.
\newblock Improving text simplification language modeling using unsimplified
  text data.
\newblock In {\em ACL},  1537--1546.

\bibitem[\protect\citeauthoryear{Kingma and Ba}{2014}]{kingma2014adam}
Kingma, D.~P., and Ba, J.
\newblock 2014.
\newblock Adam: A method for stochastic optimization.
\newblock {\em arXiv preprint arXiv:1412.6980}.

\bibitem[\protect\citeauthoryear{Lample \bgroup et al\mbox.\egroup
  }{2017}]{lample2017unsupervised}
Lample, G.; Conneau, A.; Denoyer, L.; and Ranzato, M.
\newblock 2017.
\newblock Unsupervised machine translation using monolingual corpora only.
\newblock {\em arXiv preprint arXiv:1711.00043}.

\bibitem[\protect\citeauthoryear{Lin}{1999}]{Lin1999}
Lin, D.
\newblock 1999.
\newblock Automatic identification of non-compositional phrases.
\newblock In {\em ACL},  317--324.

\bibitem[\protect\citeauthoryear{Lison and
  Tiedemann}{2016}]{lison2016opensubtitles2016}
Lison, P., and Tiedemann, J.
\newblock 2016.
\newblock Opensubtitles2016: Extracting large parallel corpora from movie and
  tv subtitles.
\newblock {\em European Language Resources Association}.

\bibitem[\protect\citeauthoryear{Luong and Manning}{}]{luong2015stanford}
Luong, M.-T., and Manning, C.~D.
\newblock Stanford neural machine translation systems for spoken language
  domains.

\bibitem[\protect\citeauthoryear{Papineni \bgroup et al\mbox.\egroup
  }{2002}]{papineni2002bleu}
Papineni, K.; Roukos, S.; Ward, T.; and Zhu, W.-J.
\newblock 2002.
\newblock Bleu: a method for automatic evaluation of machine translation.
\newblock In {\em ACL},  311--318.
\newblock Association for Computational Linguistics.

\bibitem[\protect\citeauthoryear{Poncelas \bgroup et al\mbox.\egroup
  }{2018}]{poncelas2018investigating}
Poncelas, A.; Shterionov, D.; Way, A.; Wenniger, G. M. d.~B.; and Passban, P.
\newblock 2018.
\newblock Investigating backtranslation in neural machine translation.
\newblock {\em arXiv preprint arXiv:1804.06189}.

\bibitem[\protect\citeauthoryear{Poornima, Dhanalakshmi, and
  Soman}{2011}]{poornima2011rule}
Poornima, C.; Dhanalakshmi, V.; and Soman, K.
\newblock 2011.
\newblock Rule based sentence simplification for english to tamil machine
  translation system.
\newblock {\em International Journal of Computer Applications} 25(8):38--42.

\bibitem[\protect\citeauthoryear{Sag \bgroup et al\mbox.\egroup
  }{2002}]{Sag2002}
Sag, I.~A.; Baldwin, T.; Bond, F.; Copestake, A.~A.; and Flickinger, D.
\newblock 2002.
\newblock Multiword expressions: A pain in the neck for nlp.
\newblock In {\em Proceedings of the Third International Conference on
  Computational Linguistics and Intelligent Text Processing}, CICLing '02,
  1--15.

\bibitem[\protect\citeauthoryear{Saggion \bgroup et al\mbox.\egroup
  }{2015}]{Saggion:2015:MSI:2775084.2738046}
Saggion, H.; \v{S}tajner, S.; Bott, S.; Mille, S.; Rello, L.; and Drndarevic,
  B.
\newblock 2015.
\newblock Making it simplext: Implementation and evaluation of a text
  simplification system for spanish.
\newblock {\em ACM Trans. Access. Comput.} 6(4):14:1--14:36.

\bibitem[\protect\citeauthoryear{Saluja \bgroup et al\mbox.\egroup
  }{2014}]{Saluja2014}
Saluja, A.; Hassan, H.; Toutanova, K.; and Quirk, C.
\newblock 2014.
\newblock Graph-based semi-supervised learning of translation models from
  monolingual data.
\newblock In {\em ACL}.
\newblock Baltimore, Maryland: Association for Computational Linguistics.

\bibitem[\protect\citeauthoryear{Sennrich, Haddow, and
  Birch}{2016a}]{sennrich2015improving}
Sennrich, R.; Haddow, B.; and Birch, A.
\newblock 2016a.
\newblock Improving neural machine translation models with monolingual data.
\newblock In {\em ACL},  86--96.
\newblock Berlin, Germany: Association for Computational Linguistics.

\bibitem[\protect\citeauthoryear{Sennrich, Haddow, and
  Birch}{2016b}]{sennrich2015neural}
Sennrich, R.; Haddow, B.; and Birch, A.
\newblock 2016b.
\newblock Neural machine translation of rare words with subword units.
\newblock In {\em ACL},  1715--1725.
\newblock Berlin, Germany: Association for Computational Linguistics.

\bibitem[\protect\citeauthoryear{Snover \bgroup et al\mbox.\egroup
  }{2006}]{Snover2006}
Snover, M.; Dorr, B.~J.; Schwartz, R.; Micciulla, L.; and Makhoul, J.
\newblock 2006.
\newblock A study of translation edit rate with targeted human annotation.
\newblock {\em Proceedings of Association for Machine Translation in the
  Americas}  223 -- 231.

\bibitem[\protect\citeauthoryear{{\v{S}}tajner and
  Popovic}{2016}]{vstajner2016can}
{\v{S}}tajner, S., and Popovic, M.
\newblock 2016.
\newblock Can text simplification help machine translation?
\newblock In {\em Proceedings of the 19th Annual Conference of the European
  Association for Machine Translation},  230--242.

\bibitem[\protect\citeauthoryear{Sutskever, Vinyals, and
  Le}{2014}]{Sutskever2014}
Sutskever, I.; Vinyals, O.; and Le, Q.~V.
\newblock 2014.
\newblock Sequence to sequence learning with neural networks.
\newblock In Ghahramani, Z.; Welling, M.; Cortes, C.; Lawrence, N.~D.; and
  Weinberger, K.~Q., eds., {\em NeurIPS}. Curran Associates, Inc.
\newblock  3104--3112.

\bibitem[\protect\citeauthoryear{Vaswani \bgroup et al\mbox.\egroup
  }{2017}]{Vaswani2017AttentionIA}
Vaswani, A.; Shazeer, N.; Parmar, N.; Uszkoreit, J.; Jones, L.; Gomez, A.~N.;
  Kaiser, L.; and Polosukhin, I.
\newblock 2017.
\newblock Attention is all you need.
\newblock In {\em NeurIPS}.

\bibitem[\protect\citeauthoryear{Woodsend and
  Lapata}{2011}]{Woodsend:2011:LSS:2145432.2145480}
Woodsend, K., and Lapata, M.
\newblock 2011.
\newblock Learning to simplify sentences with quasi-synchronous grammar and
  integer programming.
\newblock In {\em EMNLP},  409--420.
\newblock Stroudsburg, PA, USA: Association for Computational Linguistics.

\bibitem[\protect\citeauthoryear{Wu and others}{2016}]{wu2016google}
Wu, Y., et~al.
\newblock 2016.
\newblock Google's neural machine translation system: Bridging the gap between
  human and machine translation.
\newblock {\em arXiv preprint arXiv:1609.08144}.

\bibitem[\protect\citeauthoryear{Wubben, van~den Bosch, and
  Krahmer}{2012}]{Wubben:2012:SSM:2390524.2390660}
Wubben, S.; van~den Bosch, A.; and Krahmer, E.
\newblock 2012.
\newblock Sentence simplification by monolingual machine translation.
\newblock In {\em ACL}, ACL '12,  1015--1024.
\newblock Stroudsburg, PA, USA: Association for Computational Linguistics.

\bibitem[\protect\citeauthoryear{Xu \bgroup et al\mbox.\egroup
  }{2016}]{xu2016optimizing}
Xu, W.; Napoles, C.; Pavlick, E.; Chen, Q.; and Callison-Burch, C.
\newblock 2016.
\newblock Optimizing statistical machine translation for text simplification.
\newblock {\em Transactions of the Association for Computational Linguistics}
  4:401--415.

\bibitem[\protect\citeauthoryear{Zhang and Lapata}{2017}]{zhang2017sentence}
Zhang, X., and Lapata, M.
\newblock 2017.
\newblock Sentence simplification with deep reinforcement learning.
\newblock In {\em EMNLP},  584--594.
\newblock Copenhagen, Denmark: Association for Computational Linguistics.

\bibitem[\protect\citeauthoryear{Zhang and
  Toral}{2019}]{DBLP:journals/corr/abs-1906-08069}
Zhang, M., and Toral, A.
\newblock 2019.
\newblock The effect of translationese in machine translation test sets.
\newblock {\em CoRR} abs/1906.08069.

\bibitem[\protect\citeauthoryear{Zhao \bgroup et al\mbox.\egroup
  }{2018}]{zhao2018integrating}
Zhao, S.; Meng, R.; He, D.; Saptono, A.; and Parmanto, B.
\newblock 2018.
\newblock Integrating transformer and paraphrase rules for sentence
  simplification.
\newblock In {\em EMNLP},  3164--3173.
\newblock Brussels, Belgium: Association for Computational Linguistics.

\bibitem[\protect\citeauthoryear{Zhu, Bernhard, and
  Gurevych}{2010}]{Zhu:2010:MTT:1873781.1873933}
Zhu, Z.; Bernhard, D.; and Gurevych, I.
\newblock 2010.
\newblock A monolingual tree-based translation model for sentence
  simplification.
\newblock In {\em Proceedings of the 23rd International Conference on
  Computational Linguistics}, COLING '10,  1353--1361.
\newblock Stroudsburg, PA, USA: Association for Computational Linguistics.

\bibitem[\protect\citeauthoryear{Zoph \bgroup et al\mbox.\egroup
  }{2016}]{zoph2016transfer}
Zoph, B.; Yuret, D.; May, J.; and Knight, K.
\newblock 2016.
\newblock Transfer learning for low-resource neural machine translation.
\newblock In {\em EMNLP},  1568--1575.
\newblock Austin, Texas: Association for Computational Linguistics.

\end{thebibliography}

\end{document}